# SlowFastVAD: Video Anomaly Detection via Integrating Simple Detector and RAG-Enhanced Vision-Language Model

Zongcan Ding*
Northwestern Polytechnical
University
Xi'an, China
dingzongcan@mail.nwpu.edu.cn

Haodong Zhang*
Northwestern Polytechnical
University
Xi'an, China
hdzhang@mail.nwpu.edu.cn

Peng Wu†
Northwestern Polytechnical
University
Xi'an, China
xdwupeng@gmail.com

Guansong Pang
Singapore Management
University
Singapore, Singapore
gspang@smu.edu.sg

Zhiwei Yang
Xidian University
Guangzhou, China
zwyang97@163.com

Peng Wang
Northwestern Polytechnical
University
Xi'an, China
peng.wang@nwpu.edu.cn

Yanning Zhang
Northwestern Polytechnical
University
Xi'an, China
ynzhang@nwpu.edu.cn

## Abstract

Video anomaly detection (VAD) aims to identify unexpected events in videos and has wide applications in safety-critical domains. While semi-supervised methods trained on only normal samples have gained traction, they often suffer from high false alarm rates and poor interpretability. Recently, vision-language models (VLMs) have demonstrated strong multimodal reasoning capabilities, offering new opportunities for explainable anomaly detection. However, their high computational cost and lack of domain adaptation hinder real-time deployment and reliability. Inspired by dual complementary pathways in human visual perception, we propose **SlowFastVAD**, a hybrid framework that integrates a fast anomaly detector with a slow anomaly detector (namely a retrieval augmented generation (RAG) enhanced VLM), to address these limitations. Specifically, the fast detector first provides coarse anomaly confidence scores, and only a small subset of ambiguous segments, rather than the entire video, is further analyzed by the slower yet more interpretable VLM for elaborate detection and reasoning. Furthermore, to adapt VLMs to domain-specific VAD scenarios, we construct a knowledge base including normal patterns based on few normal samples and abnormal patterns inferred by VLMs. During inference, relevant patterns are retrieved and used to augment prompts for anomaly reasoning. Finally, we smoothly fuse the anomaly confidence of fast and slow detectors to enhance robustness of anomaly detection. Extensive experiments on four benchmarks demonstrate that SlowFastVAD effectively combines the strengths of both fast and slow detectors, and achieves remarkable detection accuracy and interpretability with significantly reduced computational overhead, making it well-suited for real-world VAD applications with high reliability requirements. The code will be released upon acceptance.

*These authors contributed equally to this work.
†Corresponding author.

## 1 Introduction

Video Anomaly Detection (VAD) aims to automatically identify abnormal events in video streams that deviate significantly from typical normal patterns. It plays a vital role in a wide range of real-world applications [55, 65, 67]. Given the rarity and high acquisition cost of anomalous samples in real-world scenarios, increasing attention has been directed toward the semi-supervised VAD paradigm, which trains models exclusively on normal videos [30, 34, 75]. By learning the underlying distribution of normal patterns, these methods attempt to detect anomalies as deviations from expected behaviors during inference.

However, semi-supervised VAD methods suffer from several inherent limitations. Since these models are trained exclusively on normal samples, they are prone to misclassifying rare yet plausible normal behaviors as anomalies, leading to high false positive rates. Moreover, existing one-class detection approaches, such as those based on autoencoders [7, 18, 40, 48, 53, 73, 83], generative adversarial networks (GANs) [13, 19, 77], or diffusion models [10, 15, 70], often exhibit limited adaptability when deployed in complex and dynamic real-world environments. In addition, most of these methods rely on end-to-end deep neural networks that are trained to fit only the distribution of normal behaviors. As a result, their decision-making results are often monotonous and lack interpretability or reasoning, making them ill-suited for scenarios where transparency and explainability are crucial.

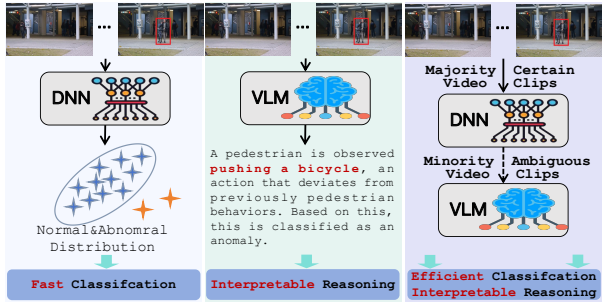

**Figure 1: Comparative analysis between conventional fast detector based on DNN (Left), recent slow detectors based on VLMs (Middle), and our SlowFastVAD (Right).**

Recently, vision-language models (VLMs) have achieved remarkable progress across various domains and have demonstrated great potential in the semi-supervised VAD task [71]. Benefiting from their multimodal integration and semantic understanding abilities, VLMs can effectively uncover latent behavioral patterns within video data. For instance, in the widely-used pedestrian street dataset Ped2, VLMs can infer the normative pattern that "only pedestrians are allowed to walk on the sidewalk" by learning from normal videos. In complex real-world scenarios, such models are capable of learning and constructing semantic representations of normal patterns, thereby enabling more accurate identification of anomalies that deviate from these learned norms. Furthermore, by leveraging their language generation capabilities and the established semantic rules, VLMs can provide clear reasoning for their detection results, significantly enhancing the interpretability and trustworthiness.

Despite the promising potential of VLMs in the VAD task, their practical application still faces several critical challenges that warrant further investigation. First, VLMs are susceptible to the hallucination, where the generated reasoning or predictions deviate from the actual video content. For example, in Ped2, VLMs occasionally misinterpret normal pedestrian walking as a crowd gathering, thereby incorrectly labeling it as an anomalous event, resulting in semantically inconsistent judgments. Second, most current VLMs are pre-trained on general-purpose datasets, and their anomaly understanding is typically based on commonsense reasoning rather than task-specific behavioral modeling. Consequently, such models may misinterpret anomalies in specific environments due to semantic ambiguity. For instance, riding a bicycle on the sidewalk is treated as an anomalous event within the Ped2 dataset's context. However, since such behavior is often deemed acceptable in real-world scenarios, the model may fail to detect it as an anomaly. Furthermore, from a deployment perspective, VLMs often incur substantial computational overhead and exhibit slow inference speeds. Performing dense inference on every video frame is particularly impractical in scenarios requiring real-time responses, such as public safety surveillance. These limitations significantly hinder the practical utility of VLM-based VAD methods.

Inspired by the dual complementary pathways in human visual perception [12, 74], namely, a cognition-driven pathway for precise understanding (slow) and an action-driven pathway for rapid response (fast), and these pathways work in tandem to respond effectively even in extreme scenarios. This paper proposes a novel VAD framework, **SlowFastVAD**, which integrates the complementary strengths of fast and slow detectors. The goal is to achieve efficient, accurate, and interpretable anomaly detection by combining a traditional feedforward network based fast detector with a high-generalization VLM based slow detector. Specifically, to enhance the adaptability and detection performance of large models in specific scenarios, we design a retrieval augmented generation (RAG) driven anomaly reasoning module. This module guides the VLM to generate various visual descriptions from normal samples, summarizes normal patterns under the given context, and further leverages Chain-of-Thought (CoT) reasoning to infer potential abnormal patterns. These normal and abnormal patterns are structured into a knowledge base. This process of knowledge base construction requires only a small number of normal samples, eliminating the requirement for full-sample training. During inference, the model retrieves relevant behavioral rules from the knowledge base based on the language description of the current video segment and incorporates them into prompts to guide the VLM toward more targeted anomaly detection. To mitigate the high computational overhead associated with the VLM inference, we propose an entropy-based intervention detection strategy. This strategy leverages the anomaly confidence generated by the fast detector to identify video segments with high uncertainty, which are then selectively forwarded to the VLM-base slow detector for further analysis. This enables significant improvements in detection accuracy and interpretability while maintaining computational efficiency. Finally, we introduce a decision fusion mechanism that integrates the predictions from both fast and slow detectors, thereby enhancing the overall robustness of the framework. We illustrate the key differences between SlowFastVAD, traditional DNN-based fast detectors and VLM-based slow detector in Figure 1, and our SlowFastVAD effectively addresses the limitations of current fast detector, namely the limited generalization capability, poor interpretability, and high computational cost of slow detector.

The main contributions of this work are summarized as follows:
- We propose the SlowFastVAD framework, which, to our knowledge, is the first to innovatively integrate the traditional fast anomaly detector with slow yet interpretable VLM-based detector, achieving a synergy between efficiency and explainability.
- We develop a RAG-driven anomaly reasoning module, in which VLM summarizes normal and abnormal patterns during training to construct a knowledge base. This knowledge base is then dynamically retrieved during inference to enhance prompts, improving the generalization to specific VAD scenarios.
- We design an entropy-based intervention detection strategy that effectively selects video segments likely to be misclassified by the fast detector, precisely triggering the VLM inference. This strategy significantly reduces overall computational costs.
- Extensive experiments on multiple public datasets demonstrate that our proposed SlowFastVAD effectively integrates the advantages of both fast and slow detectors, achieving state-of-the-art detection performance along with interpretable outputs.

## 2 Related Work
### 2.1 Non-VLM-based Video Anomaly Detection

*2.1.1 Semi-supervised VAD.* In semi-supervised VAD, training processing relies solely on normal samples, where the model learns normal patterns and identifies deviations from these patterns during inference as anomalies. Under the current deep learning paradigm, semi-supervised VAD approaches can be broadly categorized based on the network architecture into three main types: autoencoder-based approaches, generative adversarial networks (GANs)-based approaches, and diffusion-based approaches. Autoencoder-based approaches utilize an encoder to compress input samples into low-dimensional latent representations and a decoder to reconstruct the original input from the latent space. Anomalies are detected by measuring the reconstruction error between the input and output [7, 18, 40, 48, 53, 73, 83]. GAN-based approaches consist of a generator and a discriminator. The generator learns to synthesize realistic normal samples, while the discriminator aims to distinguish between real and generated data. Test samples with low authenticity scores from the discriminator are classified as anomalies [13, 19, 77]. Diffusion-based approaches progressively generate samples from noise through a reverse diffusion process. The quality of the generated samples is then used to assess the normality, with poor reconstruction indicating potential anomalies [10, 15, 70].

*2.1.2 Weakly Supervised VAD.* Weakly-supervised VAD utilizes both normal and abnormal samples during training, but lacks precise annotations of anomalies, and only coarse video-level labels are available. Current research mainly follows two paradigms: one-stage multiple instance learning (MIL) approaches [29, 54] and two-stage self-training strategies [72, 80]. To further improve detection performance, recent efforts have explored various enhancement techniques, including temporal modeling, spatiotemporal modeling, MIL-based optimization, and feature metric learning. Specifically, temporal modeling captures sequential dependencies in videos, enabling the model to utilize contextual information [11, 17, 54, 86]. Spatiotemporal modeling integrates spatial and temporal features to localize anomalous regions while suppressing background noise [26, 54]. MIL-based optimization strategies address the limitation of conventional MIL methods that focus only on high-scoring segments, by incorporating external priors, such as textual knowledge, to improve anomaly localization [9, 36]. Feature metric learning constructs a discriminative embedding space by clustering similar features and separating dissimilar ones, thereby enhancing the representation discrimination [14].

### 2.2 VLM-based Video Anomaly Detection

*2.2.1 Semi-supervised VAD.* In the field of VAD, VLMs have demonstrated significant potential and adaptability. Yang et al. proposed AnomalyRuler [71], which detects anomalies by integrating the inductive summarization and deductive reasoning capabilities of VLMs. Specifically, in the inductive phase, the model derives behavioral rules from a small number of normal samples, while in the deductive phase, it identifies anomalous frames based on these rules. In addition, Jiang et al. introduced the VLAVAD framework [25], which employs cross-modal pre-trained models and leverages the reasoning capabilities of large language models (LLMs) to enhance the interpretability and effectiveness of VAD. However, due to the slow inference speed of VLMs, the overall processing time of these methods remains high. In contrast, our SlowFastVAD integrates conventional fast detectors with VLMs, enabling sparse yet deeper reasoning based on the initial outputs of the fast detector. This design effectively balances inference speed and detection accuracy.

*2.2.2 Weakly Supervised VAD.* VLMs have also been widely applied in weakly supervised VAD. They not only enhance anomaly detection performance through visual-language enhanced features (e.g., CLIP-TSA [21]) and cross-modal semantic alignment (e.g., VadCLIP [69], TPWNG [72], and STPrompts [68]), but also contribute to interpretability by generating descriptions for anomalous events, as demonstrated in the Holmes-VAU [82]. Moreover, VLMs can be leveraged for training-free anomaly detection by utilizing their extensive prior knowledge [35, 79], offering advantages in rapid deployment and reduced computational cost. For instance, Zanella et al. [79] adopted an explainable approach in which reflective questions are used to guide the model in generating anomaly scores, without requiring additional model training.

### 2.3 VLM-based Vision Tasks

Currently, VLMs have made significant progress and found widespread application in various vision fields [57]. In image classification, VLM enhances zero-shot classification capabilities, especially in handling unknown object categories, showing excellent performance and supporting stronger domain generalization [1, 22]. In semantic segmentation, VLM improves the ability to handle unseen categories significantly by combining open-vocabulary techniques with image-text fusion [33, 60]. In video generation, VLM is used to generate consistent and multi-scene video content, pushing forward the advancement of video generation technology [31]. In cross-modal retrieval, VLM improves the effectiveness and efficiency by integrating image and language information [6, 23]. In action recognition, VLM enhances the recognition of fine-grained actions by combining pose information with language models, particularly excelling in action anticipation[39, 81].

## 3 Methodology
### 3.1 Overview

Our proposed method is illustrated in Figure 2, which consists of two branches: a fast DNN-based detector and a slow VLM-based detector. The fast detector is built upon an autoencoder-based architecture, offering high detection speed but limited interpretability. In contrast, the slow detector leverages VLMs, which provides strong interpretability at the cost of slower inference. By integrating multiple specialized components, our framework effectively combines the advantages of both detectors to achieve a balanced trade-off between efficiency and accuracy. The overall pipeline is as follows: The fast detector first performs preliminary detection and identifies potentially ambiguous segments, which are then passed to the slow detector for further analysis. The slow detector generates both anomaly confidence scores and interpretable descriptions. To select ambiguous segments more effectively, we propose an intervention detection strategy based on entropy measures. Additionally, to improve the adaptability of the VLM in specific anomaly detection scenarios, we introduce an anomaly-oriented RAG module. This module constructs a knowledge base by extracting normal patterns from training videos and inferring potential abnormal patterns, thus enhancing scene-specific reasoning capabilities. Finally, an integration mechanism combines the outputs from both detectors to yield the final prediction. This mechanism mitigates hallucination effects commonly associated with VLMs and enables the system to achieve high detection accuracy, faster inference, and interpretable output.

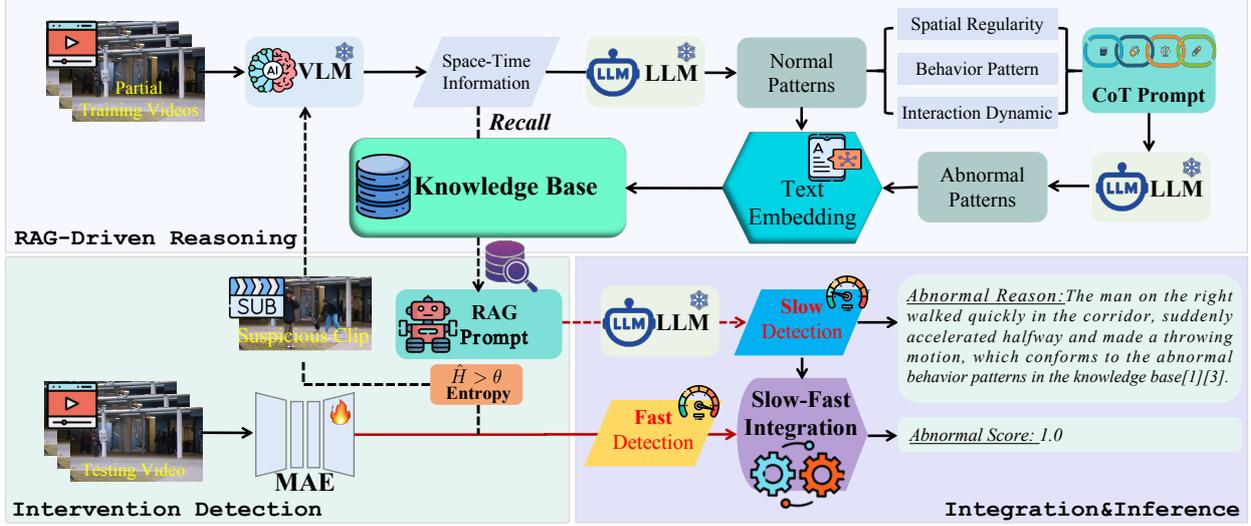

Figure 2: Overview of the proposed SlowFastVAD method. It consists of two branches: a fast DNN-based detector and a slow VLM-based detector. To seamlessly integrate the two detection branches and leverage their respective strengths, we designed three key components, i.e., intervention detection strategy, RAG-driven anomaly reasoning module, and integration mechanism, enabling an efficient and interpretable VAD framework.

### 3.2 Fast Detector

*3.2.1 Foundation Model.* In the fast detector, we adopt the AED-MAE [48], which utilizes a lightweight masked autoencoder architecture. By incorporating motion gradient based weighting, self-distillation training, and synthetic anomaly augmentation strategies, this method achieves fairly efficient VAD. AED-MAE is characterized by its compact model size and extremely fast inference speed, reaching up to 1655 FPS (frame per second).

*3.2.2 Intervention Detection Strategy.* In the context of VAD, video frames that are easy to classify typically exhibit low variance in anomaly confidence scores, resulting in low uncertainty, i.e., low entropy. However, since the fast detector is trained solely on normal samples via reconstruction, it may produce high reconstruction errors for normal-but-rare samples during inference, leading to noisy or fluctuating anomaly confidence scores. Besides, in complex scenes where the test data deviates from distributions of the training set, the fast detector may fail to generalize effectively, again causing instability in anomaly confidence scores. These fluctuations are reflected as increased entropy in the anomaly confidence scores.

To address this, we propose a novel entropy-based intervention detection strategy to identify and select ambiguous segments that are difficult to accurately classify. Specifically, given a testing video, we take its frame-level anomaly confidence scores $S_{fast}$ from the fast detector as input and partition it into a set of non-overlapping subsequences $S = \{S_q\}_{q=1}$ using a window size $n$. For each subsequence $S_q$, we compute its entropy. To account for temporal context, we apply a Gaussian filter for smoothing, integrating the entropy values of neighboring subsequences to obtain a context-aware entropy score. Given that the anomaly confidence scores are decimals ranging from 0 to 1, we adopt the differential entropy formula for calculation. The detailed calculation procedure is shown as follows.

We first estimate the probability density function of the obtained subsequence $S_q = \{score_i\}_{i=1}^n$, where $score_i$ indicates the anomaly score of the $i$-th video frame. Here, we employ the frequency distribution histogram to serve as an approximation of the probability density function for the subsequence $S_q$. The following are the detailed steps: Firstly, determine the number of histogram bins as $N_c$. Subsequently, calculate the difference between the maximum and minimum values within $S_q$. Divide the obtained difference by the number of groups $N_c$ to derive the class interval, based on which the grouping intervals can be further ascertained. On this foundation, count the number of elements of $S_q$ within each grouping interval, and then compute the corresponding frequencies to obtain the frequency distribution histogram $pdf \in \mathcal{R}^{N_c}$. For each value $score_i$ in $S_q$, first identify the group to which it belongs in the frequency distribution histogram $pdf$, and take the frequency of that group as the probability of its occurrence. In this way, the final probability density function $\hat{p}(score_i)$ of $S_q$ is obtained. Based on the obtained probability density function $\hat{p}(score_i)$ of the subsequence $S_q$, we compute the differential entropy $H_q$ of $S_q$ as follows:

$$H_q = -\sum_{i=1}^{n} \hat{p}(score_i) \log_2(\hat{p}(score_i)) \quad (1)$$

We further apply a Gaussian filter $G(\cdot)$ to $H_q$, integrating the information from neighboring subsequences $H_{q+j}$, so as to obtain the final entropy value $\hat{H}_q$ of $S_q$, which is shown below:

$$G(x) = \frac{1}{\sqrt{2\pi\sigma^2}} \exp\left(-\frac{x^2}{2\sigma^2}\right)$$
$$\hat{H}_q = \sum_{j=-Z}^{Z} H_{q+j} \cdot G(j), \quad Z = [3\sigma] \quad (2)$$

We set a threshold $\theta$ to determine which subsequences are considered uncertain. If the entropy value of a certain subsequence exceeds $\theta$, then the corresponding video segment $VF = \{frame_i\}_{i=1}^{n}$ will be fed into the slow detector for further analysis.

Moreover, to improve the interpretability of overall detection results, we also introduce a periodic sampling mechanism. Specifically, one video segment is sampled from every $T$ video segments and sent to the slow detector for semantic description and anomaly scoring. These results serve as global context cues that complement the final decision-making process with interpretable outputs.

### 3.3 Slow Detector

*3.3.1 Basic Procedure.* The input to the slow detector is the ambiguous video segment $VF$ identified by the intervention detection strategy. In the VAD task, spatiotemporal information is of crucial importance [63, 85]. Temporal information can capture the sequential evolution process of events and their durations, which helps to distinguish between normal and abnormal behaviors, because anomalies often manifest as sudden interruptions in the temporal dimension. Spatial information is divided into two parts: the foreground and the background. Foreground information focuses on the positions and motion patterns of foreground objects. Anomalies usually manifest as unusual spatial arrangements or sudden changes in positions. Background information focuses on the relatively stable scene characteristics. By understanding the background information, VLM and LLM can better extract and summarize the normal and abnormal patterns in the current scene. Based on this, $VF$ is concatenated with the CoT prompt (*refer to Appendix for details*) and then fed into the VLM (denoted as $F_{VLM}$) to extract its spatiotemporal representation $ST_{test}$. Subsequently, the spatiotemporal representation $ST_{test}$ is encoded into a vector $E$ by the embedding model *text-embedding-v2*[1] (denoted as $F_{emb}$). Detailed processes can be presented as follows:

$$ST_{test} = F_{VLM}(VF)$$
$$E = F_{emb}(ST_{test}) \quad (3)$$

Based on the similarity between $E$ and constructed patterns, the top $K$ relevant patterns $P$ and their associated binary anomaly predictions $y$ (i.e., normal and abnormal) are retrieved from the constructed knowledge base $\mathcal{D}$, which is introduced in the following section. Combine $P$ and $y$ to obtain the knowledge $J =$ $\{(P_1, y_1), \cdots, (P_K, y_K)\}$ related to the current video.

$$J = TopK(\{(P_i, y_i, sim(E, P_i))) | (P_i, y_i) \in \mathcal{D}\}) \quad (4)$$

where $sim(\cdot)$ denotes the similarity computation.

Finally, the extracted spatiotemporal representation $ST_{test}$ and the retrieved knowledge $J$ are concatenated and combined with a CoT reasoning prompt to form a structured prompt $P_{RAG} = [ST_{test}; J]$. This prompt is fed into LLM for step-by-step reasoning, producing anomaly scores $S_{slow}$ along with corresponding interpretive descriptions $R$.

$$(S_{slow}, R) = F_{VLM}(P_{RAG}) \quad (5)$$

*3.3.2 RAG-driven Anomaly Reasoning.* This module is designed to extract normal patterns from training videos, enabling VLM trained on general scenarios to better adapt to the specific VAD task. To achieve this, we apply a sparse temporal sampling strategy [59], where a segment containing $n$ consecutive frames is randomly selected from fixed-length segments of training videos. Throughout this process, we extensively incorporate the CoT prompt to guide the reasoning of models in a more interpretable and coherent manner. The overall procedure consists of four stages: visual description generation, pattern extraction and prediction, pattern refinement and aggregation, and knowledge base construction.

**Visual Description Generation**: Here, we follow the same procedure described in Section 3.3.1 to extract the spatiotemporal representation $ST_{ref}$ for the video segment.

**Pattern Extraction and Prediction:** Based on the extracted spatiotemporal representations $ST_{ref}$, we further employ the CoT prompt to guide the LLM in refining representative normal patterns $N$ (e.g., "a person walking slowly on the road" or "a small group engaged in conversation"). Building upon these patterns, the model is further prompted to reason about spatial regularity, behavioral pattern, and interaction dynamic, thereby enabling the prediction of potential abnormal patterns $A$. This step not only encodes prior knowledge of normalcy but also enhances semantic interpretability of potential anomalies. The detailed processes are presented as follows:

$$N = CoT\text{-}LLM(ST_{ref})$$
$$A = CoT\text{-}LLM(N) \quad (6)$$

where $CoT$-$LLM$ denotes the reasoning of LLMs with the assist of CoT prompt.

**Pattern Refinement and Aggregation:** After obtaining the initially extracted normal and abnormal patterns, we design a voting-based strategy for pattern refinement and aggregation. Considering that normal patterns within the same video scene often exhibit high consistency, while abnormal patterns tend to be more diverse, we aggregate highly similar patterns to refine stable behavioral representations. Meanwhile, dissimilar patterns are retained to preserve behavioral diversity. This process results in a pattern set that is both representative and diverse, laying a solid foundation for subsequent knowledge base construction. Specifically, we process the patterns summarized from the videos within each scene separately. Here, we take the normal patterns as an example for illustration, and the abnormal patterns are processed in the same way. For the $i$-th scene, the $k$-th normal pattern $N_{ik}$ is first compared for similarity with the existing patterns $N_{il}$ in the knowledge base. If the average similarity between it and the existing patterns is below the threshold $\tau$, it indicates that this pattern is dissimilar to the existing patterns in the knowledge base, and it will then be directly added to the knowledge base. Conversely, if the sum of similarities is not less than $\tau$, we identify the first $j$ normal patterns $N_{ij}$ in the knowledge base that are similar to it. These similar patterns are then aggregated and cleaned, and the aggregated and cleaned patterns are added to the knowledge base. Through continuous loop processing, after traversing all normal patterns of the $i$-th scene, we finally obtain the set $\mathcal{N}_i^*$ of all processed normal patterns for the $i$-th scene. After obtaining the set $\mathcal{N}_i^*$ of normal patterns and the set $\mathcal{A}_i^*$ of abnormal patterns for the $i$-th scene, we combine the two to obtain the set $\mathcal{P}_i$ of all patterns for this scene. The formula is expressed as follows:

$$\mathcal{N}_i^* = \bigcup_{k=1} \begin{cases} \{N_{ik}\}, \text{if } \frac{1}{|\{N_{il}\}_{l\neq k}|} \sum_{j\neq k} sim(N_{ik}, N_{il}) < \tau \\ \bigcup_{j:sim(N_{ik},N_{ij})\geq \tau}\{N_{ij}\}, \text{otherwise} \end{cases} \quad (7)$$

$$\mathcal{A}_i^* = \bigcup_{m=1} \begin{cases} \{A_{im}\}, \text{if } \frac{1}{|\{A_{in}\}_{n\neq m}|} \sum_{n\neq m} sim(A_{im}, A_{in}) < \tau \\ \bigcup_{n:sim(A_{im},A_{in})\geq \tau}\{A_{in}\}, \text{otherwise} \end{cases} \quad (8)$$

$$\mathcal{P}_i = \mathcal{N}_i^* \cup \mathcal{A}_i^* \quad (9)$$

---
[1]https://help.aliyun.com/zh/model-studio/user-guide/embedding

**Knowledge Base Construction:** The cleaned normal and abnormal patterns $\mathcal{P}_i$, along with their corresponding anomaly predictions $y_i$, are structured into standardized data formats are then encoded into vector representations using the *text-embedding-v2*[1] model, thereby constructing the knowledge base tailored for the VAD task. Mathematically, the knowledge base $\mathcal{D}$ can be expressed as follows:

$$\mathcal{D} = \bigcup_{i=1} \{F_{emb}(\mathcal{P}_i, y_i)\} \quad (10)$$

### 3.4 Slow-Fast Integration and Inference

To derive the final anomaly confidence score, we integrate $S_{fast}$ from the fast detector and $S_{slow}$ from the slow detector via a integration mechanism. First, we use the weighted-averaging method to obtain the initial fused $S_{fusion}$, which is shown as follows:

$$S_{fusion} = \alpha S_{slow} + (1-\alpha) S_{fast} \quad (11)$$

where the weighting factor $\alpha$ serves to balance the performance of fast and slow detectors. Subsequently, a Gaussian filter is applied for smoothing. Moreover, the anomaly reasoning $R$ is generated by the slow detector, endowing the detection result with high interpretability.

## 4 Experiments
## 4.1 Datasets and Evaluation Metrics

*4.1.1 Datasets.* We evaluate the proposed method on four public datasets: UCSD Ped2 [38], Avenue [32], ShanghaiTech [42], and UBnormal [61]. **UCSD Ped2** is a single-scene dataset captured on a pedestrian walkway that contains anomalies such as cyclists, skateboarders, and cars. **Avenue** is also a single-scene dataset, recorded on the main avenue of the CUHK campus, with anomalies including running and bicycling. **ShanghaiTech** is a more challenging multi-scene dataset from 13 different campus environments, characterized by variations in lighting conditions and camera perspectives. As the largest dataset for semi-supervised VAD, it comprises 270000 frames for training and approximately 50000 for testing. **UBnormal** is an open-set dataset comprising 29 synthetic scenes, where the sets of anomaly types in the training and testing splits are disjoint. For each dataset, we adopt the default training and testing splits under the semi-supervised setting, using only normal samples during training. The normal reference frames used by SlowFastVAD are randomly and uniformly sampled from normal training videos.

*4.1.2 Evaluation Metrics.* We follow recent related works [48, 71] and report the frame-level Area Under the Curve (AUC) of the Receiver Operating Characteristic (ROC). Specifically, we compute both the Micro AUC and Macro AUC. For Micro AUC, all test frames from every video are merged into a single sequence, and AUC is calculated across the entire set. In contrast, Macro AUC is computed by first calculating the AUC for each individual test video, followed by averaging these scores to obtain the final result.

### 4.2 Implementation Details

Our method is implemented using the PyTorch framework. Unless otherwise specified, Qwen-VL-Max [3] is used as VLM for visual perception, while Qwen-Max [56] serves as LLM for spatiotemporal information aggregation and retrieval-augmented generation. For Qwen-VL-Max, the sampling temperature is set to 0.01, while for Qwen-Max, the sampling temperature is set to 1.1 during mode training and 0.7 during model testing. The default hyperparameter settings for SlowFastVAD are as follows: the window size $n$ of video segment is set to 8, and the random/uniform sampling interval $T$ is set to 20 during training and reduced to 10 during testing to ensure finer temporal resolution for inference. $K$ in Eq (4) is set to 6, and the weighting factor $\alpha$ in Eq (11) is empirically set to 0.8, 0.5, and 0.7 for Ped2, Avenue, and ShanghaiTech, respectively, to balance the fast and slow detectors. For the fast detector, we follow the configuration used in AED-MAE [48].

### 4.3 Comparison with State-of-the-art Methods

In this section, we compare the proposed SlowFastVAD with dozens of baseline VAD methods across four datasets to evaluate its detection performance. Notably, for a fair comparison, we restrict our evaluation to frame- or cube-centric methods, as object-centric methods completely remove background information and irrelevant content. As shown in Tables 1 and 2, SlowFastVAD achieves overall state-of-the-art results, particularly excelling on UCSD Ped2 and UBnormal datasets, with Micro AUC scores of 99.1% and 72.2%, respectively. These results demonstrate the strong generalization ability and detection accuracy across diverse scenarios. The key advantage of SlowFastVAD lies in its dual-branch (slow and fast) architecture, which fully leverages the ability of VLMs to refine and amend the initial predictions from the fast detector. This design achieves a balanced trade-off between inference efficiency and detection accuracy. Compared to traditional VAD approaches based on visual features and reconstruction costs, SlowFastVAD benefits from the semantic understanding and external prior knowledge provided by VLMs, enabling more robust anomaly detection. For instance, compared to previous best counterpart AED-MAE [48], our SlowFastVAD yields considerable gains across different evaluation metrics. Furthermore, in contrast to VLM-only methods AnomalyRuler [71], SlowFastVAD not only achieves significantly faster inference, but also delivers improved detection performance.

Table 1: AUC scores of several state-of-the-art methods versus SlowFastVAD on Ped2 and Avenue datasets. The top three methods are shown in red, green, and blue.

| Method | Reference | Year | Ped2 AUC (%) | | Avenue AUC (%) | |
|---|---|---|---|---|---|---|
| | | | Micro | Macro | Micro | Macro |
| LSHF [84] | PR | 2016 | 91.0 | - | - | - |
| AnomalyGAN [47] | ICIP | 2017 | 93.5 | - | - | - |
| FuturePred [27] | CVPR | 2018 | 95.4 | - | 85.1 | 81.7 |
| MC2ST [28] | BMVC | 2018 | 87.5 | - | 84.4 | - |
| DeepMIL [50] | CVPR | 2018 | - | - | - | - |
| PnP-CMA [46] | WACV | 2018 | 88.4 | - | - | - |
| MemAE [16] | ICCV | 2019 | 94.1 | - | 83.3 | - |
| NNC [20] | WACV | 2019 | - | - | 88.9 | - |
| BMAN [24] | TIP | 2019 | 96.6 | - | 90.0 | - |
| AMCVAD [41] | ICCV | 2019 | 96.2 | - | 86.9 | - |
| DeepOC [66] | TNNLS | 2019 | 96.9 | - | 86.6 | - |
| StreetScene [45] | WACV | 2020 | 88.3 | - | 72.0 | - |
| MNAD [44] | CVPR | 2020 | 97.0 | - | 82.8 | 86.8 |
| SCRD [51] | ACMMM | 2020 | - | - | 89.6 | - |
| CAC [62] | ACMMM | 2020 | - | - | 87.0 | - |
| VEC-AM [75] | ACMMM | 2020 | 97.3 | - | 90.2 | - |
| AEP [76] | TNNLS | 2021 | 97.3 | - | 90.2 | - |
| LNRA [2] | BMVC | 2021 | 96.5 | - | 87.1 | - |
| TimeSformer [5] | ICML | 2021 | - | - | - | - |
| SSPCAB [49]+[27] | CVPR | 2022 | - | - | 87.3 | 84.5 |
| SSPCAB [49]+[44] | CVPR | 2022 | - | - | 84.8 | 88.6 |
| GCL [78] | CVPR | 2022 | - | - | - | - |
| FastAno [43] | WACV | 2022 | 96.3 | - | 85.3 | - |
| S3R [64] | ECCV | 2022 | - | - | - | - |
| HSNBM [4] | ACMMM | 2022 | 95.2 | - | 91.6 | - |
| ERVAD [52] | ACMMM | 2022 | 97.1 | - | 92.7 | - |
| DM-UVAD [58] | ICIP | 2023 | - | - | - | - |
| FPDM [70] | ICCV | 2023 | - | - | 90.1 | - |
| SSMCTB [37]+[27] | TPAMI | 2023 | - | - | 89.1 | 84.8 |
| SSMCTB [37]+[44] | TPAMI | 2023 | - | - | 86.4 | 86.3 |
| AnomalyRuler [71] | ECCV | 2024 | 97.9 | - | 89.7 | - |
| AED-MAE [48] | CVPR | 2024 | 95.4 | 98.4 | 91.3 | 90.9 |
| SSAE [8] | TPAMI | 2024 | - | - | 90.2 | - |
| **SlowFastVAD** | — | — | 99.1 | 99.7 | 89.6 | 93.2 |

Table 2: AUC scores of several state-of-the-art methods versus SlowFastVAD on ShanghaiTech and UBnormal datasets. The top three methods are shown in red, green, and blue.

| Method | Reference | Year | ShanghaiTech AUC (%) | | UBnormal AUC (%) | |
|---|---|---|---|---|---|---|
| | | | Micro | Macro | Micro | Macro |
| FuturePred [27] | CVPR | 2018 | 72.8 | 80.6 | - | - |
| MC2ST [28] | BMVC | 2018 | - | - | - | - |
| DeepMIL [50] | CVPR | 2018 | - | 76.5 | 50.3 | 76.8 |
| MemAE [16] | ICCV | 2019 | 71.2 | - | - | - |
| MNAD [44] | CVPR | 2020 | 68.3 | 79.7 | - | - |
| SCRD [51] | ACMMM | 2020 | 74.7 | - | - | - |
| CAC [62] | ACMMM | 2020 | 79.3 | - | - | - |
| VEC-AM [75] | ACMMM | 2020 | 74.8 | - | - | - |
| LNRA [2] | BMVC | 2021 | 75.9 | - | - | - |
| TimeSformer [5] | ICML | 2021 | - | - | 68.5 | 80.3 |
| SSPCAB [49]+[27] | CVPR | 2022 | 74.5 | 82.9 | - | - |
| SSPCAB [49]+[44] | CVPR | 2022 | 69.8 | 80.2 | - | - |
| GCL [78] | CVPR | 2022 | 78.9 | - | - | - |
| FastAno [43] | WACV | 2022 | 72.2 | - | - | - |
| S3R [64] | ECCV | 2022 | 80.4 | - | - | - |
| HSNBM [4] | ACMMM | 2022 | 76.5 | - | - | - |
| ERVAD [52] | ACMMM | 2022 | 79.3 | - | - | - |
| DM-UVAD [58] | ICIP | 2023 | 76.1 | - | - | - |
| FPDM [70] | ICCV | 2023 | 78.6 | - | 62.7 | - |
| SSMCTB [37]+[27] | TPAMI | 2023 | 74.6 | 83.3 | - | - |
| SSMCTB [37]+[44] | TPAMI | 2023 | 70.6 | 80.3 | - | - |
| AnomalyRuler [71] | ECCV | 2024 | 85.2 | - | 71.9 | - |
| AED-MAE [48] | CVPR | 2024 | 79.1 | 84.7 | 58.5 | 81.4 |
| SSAE [8] | TPAMI | 2024 | 80.5 | - | - | - |
| **SlowFastVAD** | — | — | 85.0 | 90.7 | 72.2 | 82.4 |

Table 3: Impact of each novel components on Ped2, Avenue, and ShanghaiTech datasets.

| Intervention | Integration | RAG | Ped2 | | Avenue | | ShanghaiTech | |
|---|---|---|---|---|---|---|---|---|
| | | | AUC (%) | | | | | |
| | | | Micro | Macro | Micro | Macro | Micro | Macro |
| ✗ | ✗ | ✗ | 87.8 | 89.6 | 80.1 | 86.1 | 76.3 | 82.3 |
| ✓ | ✗ | ✗ | 90.6 | 91.1 | 85.8 | 89.0 | 80.6 | 83.6 |
| ✓ | ✓ | ✗ | 94.3 | 97.2 | 86.1 | 88.5 | 83.9 | 88.4 |
| ✓ | ✓ | ✓ | **99.1** | **99.7** | **89.6** | **93.2** | **85.0** | **90.7** |

### 4.4 Ablation Studies

*4.4.1 Impact of Each Component.* In this section, we conduct an ablation study on different configurations of SlowFastVAD to evaluate the contribution of each component to overall VAD performance. The following configurations are considered: (1)Baseline: No additional components are used; the slow detector re-evaluates anomalies based solely on the fast detector's results under uniform sampling; (2) + Intervention: Only the intervention strategy is added; (3) +Intervention+ Integration: Both the intervention and integration components are used; (4) Full Model: All components, including the RAG module, are applied. The performance comparison is presented in Table 3. We observe that the baseline setting with uniform sampling yields relatively conservative performance, indicating its limited ability to capture the key temporal segments of anomalous events. Introducing the intervention strategy leads to consistent improvements across all four datasets, especially on Avenue and ShanghaiTech, confirming its effectiveness in guiding the model to focus on informative abnormal regions. Adding the integration mechanism further boosts performance, notably on Ped2 and ShanghaiTech, suggesting that it effectively combines the outputs of fast and slow detectors while better modeling temporal dependencies. Finally, incorporating the RAG module into the full model results in the best overall performance, with substantial gains on Ped2 and Avenue. This highlights the value of enhanced prompts generated by RAG in assisting the slow detector with more accurate anomaly reasoning. In summary, each component contributes to performance improvements to varying degrees. The final configuration consistently outperforms others across all datasets, particularly excelling on Ped2 and ShanghaiTech.

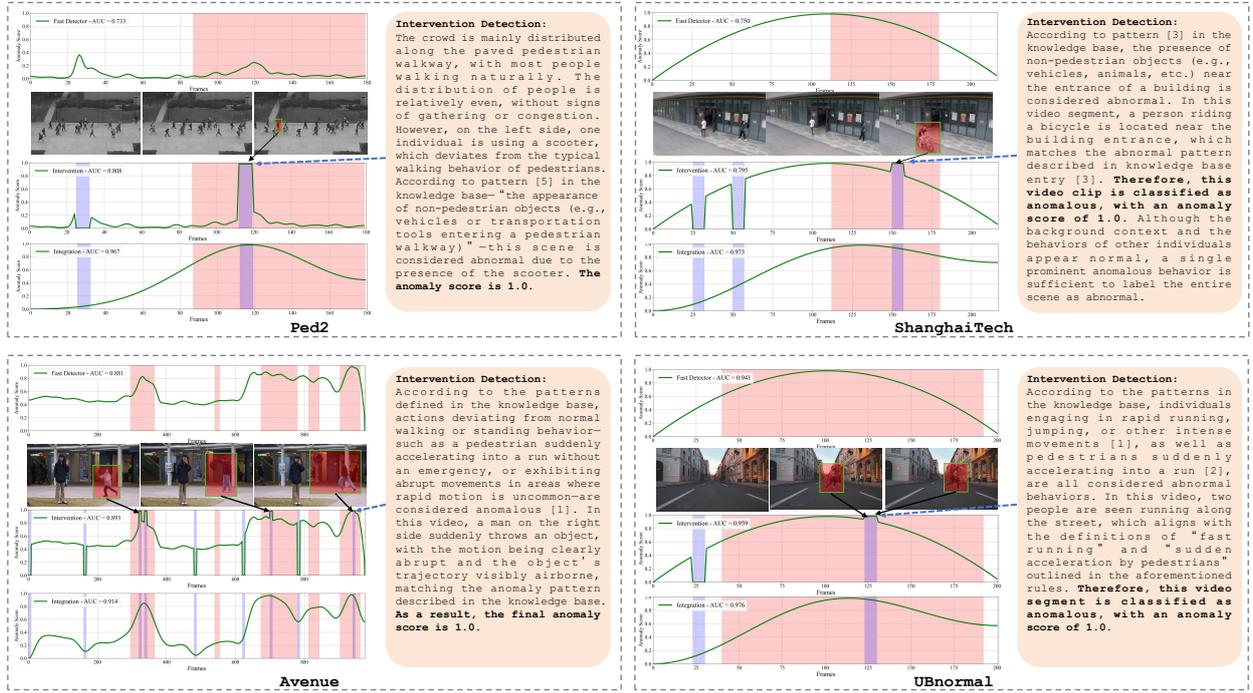

**Figure 3: Visualization of partial detection results on Ped2, Avenue, ShanghaiTech, and UBnormal. Three detection results are shown: the top displays anomaly scores generated solely by the fast detector; the middle shows the updated scores after intervention by the slow detector; the bottom presents the final results obtained through the integration of both detectors.**

**Table 4: Impact of fast detector, slow detector and the hybrid SlowFastVAD on Ped2, Avenue, and ShanghaiTech datasets.**

| Branch | Ped2 | | Avenue | | ShanghaiTech | | FPS |
|---|---|---|---|---|---|---|---|
| | AUC (%) | | | | | | |
| | Micro | Macro | Micro | Macro | Micro | Macro | |
| Fast Detector | 95.4 | 98.4 | **91.3** | 90.9 | 79.1 | 84.7 | 1655 |
| Slow Detector | 98.4 | 99.0 | 74.5 | 78.0 | **87.7** | 85.6 | 0.5 |
| SlowFastVAD | **99.1** | **99.7** | 89.6 | **93.2** | 85.0 | **90.7** | 16 |

*Note: The FPS results is obtained on a single RTX 3090 GPU. Due to limited GPU resources, the locally deployed model is Qwen2-VL-7B. If multiple GPUs are used for parallel processing, the speed can be further improved.*

*4.4.2 Impact of Different Detectors.* We further evaluated the performance of the fast detector, slow detector, and their hybrid approach across different datasets, with the results summarized in Table 4. The fast detector alone demonstrates competitive performance and delivers high inference efficiency (i.e., 1655 FPS) on all three datasets. In contrast, the slow detector exhibits relatively lower performance and considerably slow inference speed (i.e., 0.5 FPS), which can be attributed to the hallucination effects commonly observed in LLMs when operating independently, thereby compromising their ability to accurately identify anomalous events. By integrating both detectors, the hybrid approach achieves the superior overall performance across all datasets. Although a slight decrease in Micro AUC is observed on Avenue dataset, the dual-branch combination effectively suppresses hallucination effects, significantly reducing false positives and false negatives while leveraging the strengths of the fast detector. Moreover, the hybrid approach maintains a favorable balance between detection accuracy and real-time inference (16 FPS), making it a practical and robust solution for VAD in diverse scenarios. Moreover, this also substantiates the effectiveness of our biologically inspired design, which emulates the human visual system's dual complementary pathways, namely, mimicking the coordination between rapid action-oriented responses and slower cognition-driven reasoning.

### 4.5 Qualitative Analyses

Figure 3 visualizes the detection results of our SlowFastVAD and its variants on different datasets. The abnormal parts are highlighted with green bounding boxes in video frames. In the detection result, the red sections represent video segments labeled as abnormal in ground truth, while the blue sections represent the detection results after the intervention of slow detector. It is evident that using only the fast detector can achieve relatively good detection performance; However, it still suffers from noticeable false positives and false negatives, especially as observed in samples from Ped2 and Avenue. By incorporating the slow detector based on VLM through the intervention stragety to analyze suspicious regions, the local detection performance is significantly improved. Nevertheless, the localized enhancements have limited influence on the overall prediction. Therefore, the final integration of the fast and slow detectors via a Gaussian filter leads to a more globally consistent improvement, further enhancing overall detection performance.

In addition, we present several representative reasoning results from the slow detector. Due to space limitations, we randomly select a subset of intervention segments for illustration. Compared to the fast detector, which relies on simple data fitting to produce anomaly scores, the VLM-based slow detector leverages both pre-trained knowledge and domain-specific information introduced via the RAG module to enable brain-inspired deep reasoning over events, thereby producing more interpretable and accurate anomaly assessments.

## 5 Conclusion

In this work, we introduce SlowFastVAD, a novel hybrid framework that integrates a fast anomaly detector with a retrieval augmented generation enhanced vision-language model to achieve both efficiency and interpretability in video anomaly detection. The fast detector provides initial detection results, while several ambiguous segments are selectively analyzed by the slower yet more explainable VLM, reducing unnecessary computational overhead. By leveraging this dual-branch detection pipeline, our method effectively balances computational cost and detection accuracy. Specifically, the proposed entropy-based intervention strategy ensures that only uncertain segments are processed by the VLM, while the construction of a domain-adapted knowledge base further enhances the VLM's adaptability to specific VAD scenarios. Extensive experiments conducted on four datasets demonstrate that SlowFastVAD outperforms existing methods, achieving state-of-the-art detection performance while maintaining interpretability. In the future, we will further explore task-specific foundation models centered on VAD and continue to enhance reasoning efficiency.